\documentclass[letterpaper, 10 pt, conference]{ieeeconf}  % Comment this line out if you need a4paper

\IEEEoverridecommandlockouts                              % This command is only needed if 
\usepackage{graphics} % for pdf, bitmapped graphics files
\usepackage{graphicx}% http://ctan.org/pkg/graphicx
\usepackage{epsfig} % for postscript graphics files
\usepackage{mathptmx} % assumes new font selection scheme installed
\usepackage{times} % assumes new font selection scheme installed
\usepackage{amsmath,bm} % assumes amsmath package installed
\usepackage{amssymb}  % assumes amsmath package installed
\usepackage{amsmath,mathtools}
\usepackage{xcolor}
\usepackage[noadjust]{cite}
\usepackage{url}
\usepackage{lipsum}
\usepackage{flushend}

\usepackage[linesnumbered,lined,ruled]{algorithm2e}
\usepackage{algorithmic}
\SetKwInOut{Input}{input}
\SetKwInOut{Output}{output}
\SetKw{Continue}{continue}
\newcommand*\Let[2]{ #1 $\gets$ #2}

\SetCommentSty{mycommfont}

\makeatletter
\def\@citex[#1]#2{\leavevmode
\let\@citea\@empty
\@cite{\@for\@citeb:=#2\do
{\@citea\def\@citea{,\penalty\@m\ }%
\edef\@citeb{\expandafter\@firstofone\@citeb\@empty}%
\if@filesw\immediate\write\@auxout{\string\citation{\@citeb}}\fi
\@ifundefined{b@\@citeb}{\hbox{\reset@font\bfseries ?}%
\G@refundefinedtrue
\@latex@warning
{Citation `\@citeb' on page \thepage \space undefined}}%
{\@cite@ofmt{\csname b@\@citeb\endcsname}}}}{#1}}
\makeatother

\usepackage{subcaption}

\makeatletter
\let\NAT@parse\undefined
\makeatother
\usepackage[pagebackref,breaklinks,colorlinks]{hyperref}

\begin{document}

\title{\LARGE \bf
Tightly-coupled Visual-DVL-Inertial Odometry for Robot-based Ice-water Boundary Exploration

\thanks{This work is supported by the National Science Foundation (NSF) under
        the award \#1945924, and the Graduate School of Oceanography, University
        of Rhode Island. We also thank the field support from the Great Lake Research Center, Michigan Technological University}

\thanks{$^{\dagger}$The author is with the Department of Ocean Engineering, 
        University of Rhode Island, Narragansett, RI 02882, USA. Email:
        {\tt\small linzhao@uri.edu}}

\thanks{$^{\ddagger}$The authors are with the Graduate School of Oceanography,
        University of Rhode Island, Narragansett, RI 02882, USA. Emails:
        {\tt\small \{mzhou,bloose\}@uri.edu}}
}

\author{Lin Zhao$^{\dagger}$, Mingxi Zhou$^{\ddagger}$ and Brice Loose$^{\ddagger}$ }

\maketitle
\thispagestyle{empty}
\pagestyle{empty}

%%%%%%%%%%%%%%%%%%%%%%%%%%%%%%%%%%%%%%%%%%%%%%%%%%%%%%%%%%%%%%%%%%%%%%%%%%%%%%%%
%%%%%                              Abstract                                %%%%%
%%%%%%%%%%%%%%%%%%%%%%%%%%%%%%%%%%%%%%%%%%%%%%%%%%%%%%%%%%%%%%%%%%%%%%%%%%%%%%%%

% !TEX root = main.tex

%%%%%%%%%%%%%%%%%%%%%%%%%%%%%%%%%%%%%%%%%%%%%%%%%%%%%%%%%%%%%%%%%%%%%%%%%%%%%%%%
%%%%%                              Abstract                                %%%%%
%%%%%%%%%%%%%%%%%%%%%%%%%%%%%%%%%%%%%%%%%%%%%%%%%%%%%%%%%%%%%%%%%%%%%%%%%%%%%%%%

\begin{abstract}
% motivation
Underwater robots, like Autonomous Underwater Vehicles (AUVs) and Remotely Operated Vehicles (ROVs), are promising tools for the exploration and study of the under-ice environment and the ecosystems that thrive there. 
% challengings
However, state estimation is a well-known problem for robotic systems, especially, for the ones that travel underwater.
In this paper, we present a tightly-coupled multi-sensors fusion framework to increase localization accuracy that is robust to sensor failure. 
Visual images, Doppler Velocity Log (DVL), Inertial Measurement Unit (IMU) and Pressure sensor are integrated using a Multi-State Constraint Kalman Filter (MSCKF) for state estimation.
Besides, a modified keyframe-based clone marginalization and a new DVL-aided feature enhancement method are presented to further improve the localization performance.
% results.  
The proposed method is validated in the under-ice environment on Lake Michigan, USA, and the results are cross-compared with 10 other different sensor fusion setups.
Overall, the integration of keyframe enabled and DVL-aided feature enhancement yielded the best performance with a Root-mean-square error of less than 2 m compared to the ground truth path over a total traveling distance of about 200 m.
\end{abstract}

%%%%%%%%%%%%%%%%%%%%%%%%%%%%%%%%%%%%%%%%%%%%%%%%%%%%%%%%%%%%%%%%%%%%%%%%%%%%%%%%
%%%%%                              Introduction                            %%%%%
%%%%%%%%%%%%%%%%%%%%%%%%%%%%%%%%%%%%%%%%%%%%%%%%%%%%%%%%%%%%%%%%%%%%%%%%%%%%%%%%

% !TEX root = main.tex

%%%%%%%%%%%%%%%%%%%%%%%%%%%%%%%%%%%%%%%%%%%%%%%%%%%%%%%%%%%%%%%%%%%%%%%%%%%%%%%%
%%%%%                              Introduction                            %%%%%
%%%%%%%%%%%%%%%%%%%%%%%%%%%%%%%%%%%%%%%%%%%%%%%%%%%%%%%%%%%%%%%%%%%%%%%%%%%%%%%%

%%%% motivations
%%%% based on existing challenges, step by step talk about why we do this work
\section{Introduction}
\label{sec:introduction}

%% Motivation:
%%  - scientific meaning of polar region
%%  - AUV is a potential tool for ice-water interface
%%  - underwater state estimation is an open problem
The ocean in polar regions plays a vital role, affecting the global biogeochemical cycle~\cite{SeaIceRole-Vancoppenolle-2013, Glacier-VolcanicSource_Loose_2018} and the ice-water interface is an active region where transports between ice and water can set the biogeochemical and determine growth in the zones where sunlight penetrates. Despite limited resources, the under ice environment can experience large plankton blooms \cite{arrigo2012massive}, but very little is understood about how they take place.  
Previously, underwater vehicles have been successfully deployed to collect physical and biogeochemical measurements under sea ice~\cite{IceThickness_Wadhams_2012,AlgalCommunities_Robison_2011}.
Recently, under-ice AUV deployments~\citeleft\citenum{Icefin-Spears-2016, Autosub2000_Phillips_2020, Icemap-Williams-2014}\citeright ~have been carried out in regions that were previously inaccessible to observation, and to produce a higher spatial coverage to fill the observation gaps left by traditional ice-anchored instruments and ice coring.  AUVs and ROVs are uniquely capable of observing these environments in a manner that creates almost no disturbance, while traditional ice observations involve perforating the ice, which upsets the delicate physical structure, thereby biasing the resulting measurements.
However, the localization (i.e., state estimation) is particularly challenging in underwater environments due to the lack of GPS~\cite{AUV-Localization-Review_Paull_2014}.
Under-ice AUV operation is an extreme case that AUVs could not return to the surface to obtain the GPS fixes for bounding the localization drift \cite{UnderIce-VehicleDesign_Barker_2020}. 
Therefore, the collected measurements are hard to georeference, challenging the creation of spatially coherent maps of the processes, e.g., the algae patchiness, and sea-ice surface topography.

%% Challenge 1: challenging motion
%   - acoustic beacon
%   		- installation and calibration	
% 	- DVL: 
% 		- drift
% 		- more drift for low speed (SNR higher)  
Acoustic transducer arrays, such as Long Baseline (LBL) \cite{LBL-Underice_Jakuba_2008} and Ultra-short Baseline (USBL) \cite{USBL-Underice_Kukulya_2010}, are commonly used for under-ice navigation.
But, they will require installation and extrinsic calibration for transducers.
In the under-ice environment, the performance of acoustic communication devices may degrade due to up-bending sound propagation and ice keel blockage~\cite{UnderIce-AcousticCom_Freitag_2017}.
To this end, self-contained underwater navigation methods are researched and often used as it requires fewer logistic operations.
The most common one is the dead-reckoning navigation~\cite{DVL-Nav_Whitcomb_1999} which fuses velocity measured by a DVL and an inertial measurement sensor.
However, this method suffers from unbounded position errors ranging from 8 to 22 m for different quality Inertial Navigation System (INS)~\cite{ModelNav_Randeni_2018}.
% We also have observed that if an AUV is experiencing irregular motions such as low-speed cruising, quick turns and hovering, the navigation drift will grow rapidly possibly due to the reduced signal-to-noise (SNR) ratio in the IMU and DVL measurements. 

%% Challenge 2: challenging images
%%  - Visual:
%%    - dynamic illumination
%%		- motion blur
%%		- light attenuation 
%%    - limited visibility
%%  - VIO:
%% 		- more detail, workng with imu
%%		- still not stable
For the ice-water boundary exploration, vehicles would maintain a close distance (1 to 2 m) in order to collect vital measurements (such as sea-ice roughness, light penetration and water density) to study the ice-water exchanges. 
The close distance makes it possible to use camera images to aid underwater localization for AUVs.
In recent years, Visual-Odometry (VO)/Simultaneous Localization and Mapping (SLAM) has drawn increased attention, serving the rapid development in robot autonomy.
To improve the robustness of state estimation, Visual-Inertial-Odometry (VIO)~\cite{VIO-Review_Huang_2019} has been widely used. 
However, underwater visual-based SLAM is still challenging because of the dynamic illumination, limited visibility, light obstruction, texture-less area and motion blur \cite{Underwater-VIO-Survey_Joshi_2019}. 
Also, Mono-VIO has a well-known issue that the metric scale is not observable if there is no acceleration excitation. 
In such case, additional sensors (e.g., stereo-camera~\cite{SVIn2_Rahman_2019}) are needed for reliable and accurate SLAM solutions in the underwater environment.

%% Challenge 3: 
%% 	- realtime running onboard
The algorithm development always has to consider the trade-off between accuracy and computational cost when deploying on the robotics platform. 
Non-linear optimization-based VIO (e.g., OKVIS~\cite{OKVIS_Leutenegger_2015}) allows for the reduction of error through relinearization but with a high computational cost. 
Filtering-based VIO (e.g., MSCKF~\cite{MSCKF_Mourikis_2007}) is proven to be efficient and accurate in resources constraint applications, e.g, the state-of-the-art ARCore~\cite{ARCore} running on mobile devices.
To deal with the degraded performance that may exist when only using the one-time linearization, observability-based methodology~\cite{OC-MSCKF_Huang_2008} can be applied to improve the consistency. 
More details about the relevant works can be found in Section II.

%% our work DVL-Pressure-VIO
In this paper, we present a multi-modal sensor fusion framework to address the two challenges in underwater environments: degenerate motion and challenging image conditions.
Our method fuses the measurements from DVL, IMU, camera and pressure sensor in the MSCKF framework to improve the robustness of state estimation.
Our main contributions\footnote{code: \url{https://github.com/GSO-soslab/msckf_dvio}} are:
\begin{itemize}
  \item A modified keyframe marginalization method to increase the tracking period of features.
  \item A DVL point cloud enhanced feature position recovery with a new data association and estimation approach, presented in Section IV.
  \item Algorithm validation and comparison with different sensor enable/disable settings using a real-world under-ice ROV data set, presented in Section V
  % \item compare with state-of-art VIO algorithm
\end{itemize} 

%% Paper structure
The remaining paper is organized as follows: the next section reviews the related work on multi-sensor fused state estimation with an emphasis on underwater environments.
Section~\ref{sec:filter_description} introduces the filter implementation with multi-sensor setup and the implementation of the keyframe method. 
Section~\ref{sec:enhancement} presents our DVL-aided feature enhancement approach.
Section~\ref{sec:experiment} presents the experiment results, and we will conclude the paper and discuss our future plans in Section~\ref{sec:conclusions}.

%%%%%%%%%%%%%%%%%%%%%%%%%%%%%%%%%%%%%%%%%%%%%%%%%%%%%%%%%%%%%%%%%%%%%%%%%%%%%%%%
%%%%%                              Related Work                            %%%%%
%%%%%%%%%%%%%%%%%%%%%%%%%%%%%%%%%%%%%%%%%%%%%%%%%%%%%%%%%%%%%%%%%%%%%%%%%%%%%%%%

% !TEX root = main.tex

%%%%%%%%%%%%%%%%%%%%%%%%%%%%%%%%%%%%%%%%%%%%%%%%%%%%%%%%%%%%%%%%%%%%%%%%%%%%%%%%
%%%%%                              Related Work                            %%%%%
%%%%%%%%%%%%%%%%%%%%%%%%%%%%%%%%%%%%%%%%%%%%%%%%%%%%%%%%%%%%%%%%%%%%%%%%%%%%%%%%

\section{Related Work}
\label{sec:related_work}
% Underwater localization strategies can be generally divided into two types: external-aided approaches and self-contained methods.
% External-aided approaches normally use distributed acoustic arrays to locate a underwater vehicle based on the time-of-flight of acoustic signals transmitted between the vehicle and the acoustic array. 
% xxx give several examples.
% Then, go to the DVL-fused methods, then SLAM (terrain-aided navigation), and visual-aided approaches.
% (In each types, you have to have a sentence to mention the limitation in ice-water boundary environment, this section is more of a detailed discussion of the related work to support your goals and application discussed in Section I).

%%% related 1: DVL-fused(velocity, pointcloud) 
When an underwater vehicle is operated close to a target (e.g., seafloor or sea-ice), the measurements from DVL and INS are commonly fused for dead-reckoning (DR).
To increase localization accuracy, DR integrated with a pressure sensor in a tightly-coupled EKF for under-ice navigation was presented in ~\cite{INS-DVL-Pressure_Zhao_2022}. 
In addition, acoustic beacon aided DR solutions using LBL~\cite{INS-DVL-LBL_Miller_2010} or USBL~\cite{INS-DVL-USBL_Li_2023} have existed for years to bound the odometry drift.
When other acoustic sensors, such as multibeam echosounder (MBES)~\cite{INS-DVL-MBES_Palomer_2016} and multibeam forward-looking sonar (MFLS)~\cite{INS-DVL-MFLS_Cheng_2022}, are available, geophysical observations made by the sonars could further improve localization robustness.
Besides velocity measurements, DVL also generates sparse point clouds to provide additional measurements for localization. 
For example, in~\cite{DVL-Cloud-aided_Ozog_2013}, the author presented the factor-graph SLAM using parameterized planar features from sparse point clouds.

%%% related 2: visual-fused(IMU, sonar, DVL ...)
Compared to acoustic sensors, visual data contain more information and could be leveraged for localization. 
Recently, visual-based SLAM~\cite{OpenVINS_Geneva_2020,ORB-SLAM3_Campos_2021} has been significantly researched in the robotic community across various domains and applications.
Even though there is significant SLAM research done in the marine robotics community, significant technological hurdles remain, such as poor image quality in a low light environment, featureless ice terrain, and limited onboard processing capability.

Offline methods, such as the Structure From Motion (SfM), have been applied to recover large-scale underwater 3D scenes and camera poses~\cite{Underwater-SFM_Singh_2007,Mono-Mosaicing_Nicosevici_2009}. 
Yet, these methods are computationally expensive and not realistic to run online on an AUV. 
On the other hand, VO is capable of processing images at high frame rates (e.g., 10-20Hz), and the uses of monocular camera and stereo-camera have been investigated for underwater scenarios~\citeleft\citenum{Mono-VO-Mosaic_Gracias_2003,Mono-VO-Turbid_Ferrera_2019,Stereo-VO_OpticalFlow_Corke_2007,Stereo-VO-Selective_Bellavia_2017}\citeright.
As mentioned in the survey~\cite{Underwater-VIO-Survey_Joshi_2019}, visual measurements are usually combined with other sensors (e.g., IMU) for improved performance. 
For example, the pressure sensor can be used to aid the VIO algorithms using filter-based and optimization-based methods~\cite{VIPO-Filter_Shkurti_2011} and~\cite{VIPO-Optimization_Hu_2022}.  
The SVIn2~\cite{SVIn2_Rahman_2019} took another step forward, which fuses the measurements from stereo cameras, IMU, depth sensor and profiling sonar in a keyframe-based nonlinear optimization for underwater localization.
Similar to our method presented in this paper, DVL velocity fused VIO are developed in~\citeleft\citenum{DVIO-HullInspection_Kim_2013, DVIO-Habor_Vargas_2021,DVIO-DeepSea_Chavez_2019}\citeright ~for hull inspection, harbor exploration and deep-sea operation. 
However, our work herein furthers the field by using the sparse point cloud from the DVL to enhance the feature estimation for vehicle pose updates.

%%% related 3: Range based enhacement

Dense point clouds fusing with images have been widely used to enhance feature 3D position estimation in computer vision. 
In~\cite{LiDAR-SfM_Zhen_2020}, the authors presented a LiDAR-enhanced SfM pipeline that fuses the dense LiDAR point clouds with the matched visual features in a joint optimization to solve camera motion and feature position. 
In \cite{Depth-VO_Zhang_2014,LVI-SAM_Shan_2021}, LiDAR point cloud is used to interpolate the depth for the detected camera features.
However, such dense point clouds are typically not available for underwater robotics, or a power-hungry multibeam sonar is needed.
To our best knowledge, there is limited research on using sparse point clouds for Visual SLAM.
For example, the method presented in~\cite{Echosounder-VO_Roznere_2020,Depth-VIO-Underwater_Xu_2021} used the sparse range measurements from a single-beam echosounder to recover the depth information for a monocular SLAM system.
To further advance the visual SLAM using sparse point clouds, our method in this paper will employ non-uniform sparse point clouds from a DVL sensor to aid feature 3D position estimation.
%%

%%%%%%%%%%%%%%%%%%%%%%%%%%%%%%%%%%%%%%%%%%%%%%%%%%%%%%%%%%%%%%%%%%%%%%%%%%%%%%%%
%%%%%                               Method                                 %%%%%
%%%%%%%%%%%%%%%%%%%%%%%%%%%%%%%%%%%%%%%%%%%%%%%%%%%%%%%%%%%%%%%%%%%%%%%%%%%%%%%%

% !TEX root = main.tex

%%%%%%%%%%%%%%%%%%%%%%%%%%%%%%%%%%%%%%%%%%%%%%%%%%%%%%%%%%%%%%%%%%%%%%%%%%%%%%%%
%%%%%                                Method                                %%%%%
%%%%%%%%%%%%%%%%%%%%%%%%%%%%%%%%%%%%%%%%%%%%%%%%%%%%%%%%%%%%%%%%%%%%%%%%%%%%%%%%

\section{Filter Description}
\label{sec:filter_description}

In this section, we will present the tightly-coupled multi-sensors fusion based on the state-of-art MSCKF \cite{MSCKF_Mourikis_2007} framework.
For underwater robots, we further present the measurement update equations for the DVL and pressure sensor, followed by our keyframe selection strategy.

\subsection{State Vector} 
We follow the notation in~\cite{OpenVINS_Geneva_2020} and define the system state to be $\mathbf{x}_k$ at the time step, $k$.
As shown in Eq.~\ref{eq:states} to \ref{eq:states_3}, the system state consists of the current IMU state, $\mathbf{x}_{IMU}$ and the clone state, $\mathbf{x}_{C}$, which contains $n$ past IMU poses.
In Eq.~\ref{eq:states_2} and \ref{eq:states_3}, ${^{I_k}_{G}}\bar{q}$ \cite{IndirectKF-Trawny-2005} is the unit quaternion representing the rotation from the global frame $\{G\}$ to the IMU frame $\{{I_k}\}$ at time $k$,
${^G}\mathbf{p}_{I_k}$ and ${^G}\mathbf{v}_{I_k}$ are the IMU position and velocity with respect to $\{G\}$, 
$\mathbf{b}_g$ and $\mathbf{b}_a$ describe the biases of the angular velocity and linear acceleration measured by the gyro and accelerometer in an IMU. 
The cloned IMU poses are denoted by $\{{^{I_i}_{G}}\bar{q}$ and $ {^G}\mathbf{p}_{I_i}\}, i \in [1,n]$.
\vspace{-2ex}
\begin{align}
\mathbf{x}_k  &= 
        \left[ 
               \begin{array}{cc} 
                 \mathbf{x}_{IMU}^\top  & 
                 \mathbf{x}_{Clone}^\top 
               \end{array} 
        \right]^\top 
        \label{eq:states} \\
\mathbf{x}_{IMU} &= 
        \left[ 
               \begin{array}{ccccc} 
                 {^{I_k}_{G}}\bar{q}^\top  & 
                 {^G}\mathbf{p}_{I_k}^\top &
                 {^G}\mathbf{v}_{I_k}^\top &
                 \mathbf{b}_g^\top     &
                 \mathbf{b}_a^\top 
               \end{array} 
        \right]^\top 
        \label{eq:states_2} \\
\mathbf{x}_{Clone} &= 
        \left[ 
               \begin{array}{ccccc} 
                 {^{I_1}_{G}}\bar{q}^\top  & 
                 {^G}\mathbf{p}_{I_1}^\top &
                 \cdots &
                 {^{I_n}_{G}}\bar{q}^\top     &
                 {^G}\mathbf{p}_{I_n}^\top 
               \end{array} 
        \right]^\top 
        \label{eq:states_3}
\end{align}

In this paper, we define $\mathbf{x}=\hat{\mathbf{x}} \boxplus \tilde{\mathbf{x}}$, 
where $\mathbf{x}$ is the true state, $\hat{\mathbf{x}}$ is estimation, $\tilde{\mathbf{x}}$
is the error state, and the $\boxplus$ operation maps the vector to a given manifold
\cite{Manifold-Fusion_Hertzberg_2013}. For quaternions, we define the quaternion boxplus 
operation using the left quaternion error in Eq.~\ref{eq:q_manifold} where $\boldsymbol{\theta}$ is the Euler angles.
\begin{equation}
\label{eq:q_manifold}
    \bar{q} \boxplus \delta \boldsymbol{\theta} 
        \triangleq 
    \left[ \begin{array}{c} \dfrac{1}{2}\delta  \boldsymbol{\theta}  \\ 1 
           \end{array} 
    \right] \otimes \bar{q}
\end{equation}

\subsection{IMU Propagation}
The state is propagated from $k-1$ to $k$ time step using the generic nonlinear IMU kinematics model  \cite{Fundamental-Inertial-Nav_Chatfield_1997} with IMU measurements, including linear accelerations (${^I}\mathbf{a}_m$) and angular velocities (${^I}\bm{\omega}_m$).
\begin{align}\label{eq:prop_nonlinear}
    \mathbf{x}_k = f(\mathbf{x}_{k-1}, {^I}\mathbf{a}_m, {^I}\boldsymbol{\omega}_m, \mathbf{n}_I)
\end{align}
where $\mathbf{n}_I = \left[ 
                        \begin{array}{cccc} 
                            \mathbf{n}_g^\top    & 
                            \mathbf{n}_a^\top    &
                            \mathbf{n}_{\omega g}^\top &
                            \mathbf{n}_{\omega a}^\top        
                        \end{array} 
                    \right]^\top $,
including the zero-mean Gaussian noise ($\mathbf{n}_g$ and $\mathbf{n}_a$) and the random walk bias noise ($\mathbf{n}_{\omega g}$ and $\mathbf{n}_{\omega a}$) for the gyroscope and accelerometer.
The estimated state and propagated covariance are:
\begin{align}\label{eq:propagated}
    \mathbf{\hat{x}}_{k|k-1} & = 
        f(\mathbf{\hat{x}}_{k-1|k-1}, {^I}\mathbf{a}_m, {^I}\boldsymbol{\omega}_m, \mathbf{0}) \\
    \mathbf{P}_{k|k-1}       & = 
        \boldsymbol{\Phi}_{k|k-1} \mathbf{P}_{k-1|k-1} \boldsymbol{\Phi}_{k|k-1}^\top +
        \mathbf{G}_{k-1} \mathbf{Q} \mathbf{G}_{k-1}^\top
\end{align}
where $\mathbf{\hat{x}}_{k|k-1}$ denotes the estimated state at time $k$ given the measurements at time $k-1$, $\boldsymbol{\Phi}_{k|k-1}$ and $\mathbf{G}_{k-1}$ are system Jacobian and noise Jacobian of the nonlinear system \cite{MSCKF_Mourikis_2007}, and $\mathbf{Q}$ is a discrete-time covariance matrix of IMU noise $\mathbf{n}_I$.

% \subsection{State Augmentation}

% When the image keyframe is determined, IMU state will propagate forward to the timestep of the image. 
% The propagated inertial state is cloned to $\mathbf{x}_C$, and the covariance matrix is augmented using the equation below.

% \begin{align}\label{eq:cov_aug}
% \mathbf{P}_{k|k}     = \left[ 
%                           \begin{array}{c} 
%                              \mathbf{I}_{(15+6n)}       \\ 
%                              \mathbf{J}
%                           \end{array} 
%                          \right] 
%                          \mathbf{P}_{k|k}  
%                          \left[ 
%                           \begin{array}{c} 
%                              \mathbf{I}_{(15+6n)}       \\ 
%                              \mathbf{J} 
%                           \end{array} 
%                          \right]^\top                   
% \end{align}
% where the Jacobian $\mathbf{J}$ is given by:
% \begin{align}\label{eq:cov_aug_J}
% \mathbf{J} = 
%   \left[ 
%     \begin{array}{cccc} 
%        \mathbf{I}_{3\times3} & \mathbf{0}_{3\times3}  &
%        \mathbf{0}_{3\times9} & \mathbf{0}_{3\times6n} \\ 
%        \mathbf{0}_{3\times3} & \mathbf{I}_{3\times3}  &
%        \mathbf{0}_{3\times9} & \mathbf{0}_{3\times6n}
%     \end{array} 
%   \right]                 
% \end{align}

\subsection{DVL Velocity Measurement Update}

% measurement function
The DVL velocity measurement is defined in Eq.~\ref{eq:dvl_measurement} which is a function of the linear and angular velocity in the IMU frame, the relative transformation (${^I_D}\mathbf{R}$ and ${^I}\mathbf{p}_D$) between the IMU frame and the DVL frame, and the rotation (${^{I_k}_G}\mathbf{R}$) from the global frame to the IMU frame.
We also have the measurement noise $\mathbf{n}_D \thicksim \mathcal{N}(\mathbf{0}, \mathbf{R}_D)$, and the skew-symmetric matrix of the IMU's angular velocity denoted by
$\lfloor {^{I_k}}\bm{\omega} \rfloor_\times$.
For EKF update, the Jacobian matrix $\mathbf{H}_{D,k}$ with respect to the state, $\mathbf{x}_k)$ can be found in \cite{INS-DVL-Pressure_Zhao_2022}.
\begin{align}\label{eq:dvl_measurement}
    \mathbf{z}_{D,k} = h_D(\mathbf{x}_k)+ \mathbf{n}_D 
                     = {^I_D}\mathbf{R}^\top 
                     ({^{I_k}_G}\mathbf{R} {^G}\mathbf{v}_{I_k} + \lfloor {^{I_k}}\bm{\omega} \rfloor_{\times} {^I}\mathbf{p}_D)
                     + \mathbf{n}_D
\end{align}

\subsection{Pressure Measurement Update}
The pressure measurement can be written in Eq.~\ref{eq:z_measurement} where $\mathbf{s}=[\begin{array}{ccc} 0 & 0 & 1 \end{array}]$ used for selecting the third dimension, ${^P}\mathbf{P}_{in} = [\begin{array}{ccc} 0 & 0 & {^P}p_{in} \end{array}]^\top$ and ${^P}p_{in}$ are the pressure measurement at the initial position, ${^P}\mathbf{P}_{k} = [\begin{array}{ccc} 0 & 0 & {^P}p_{k} \end{array}]^\top$ and ${^P}p_k$ is the pressure measurement at timestamp $k$, the three rotation matrices (${^D_P}\mathbf{R}$, ${^I_D}\mathbf{R}$, and ${^{I_k}_G}\mathbf{R}^\top$) are used to transform the pressure measurement into the global frame, and $n_{p_z}$ is a zero-mean white Gaussian noise. 
For EKF update, the Jacobian matrix $\mathbf{H}_{p_z}$ with respect to the state, $\mathbf{x}_k)$, can be found in \cite{INS-DVL-Pressure_Zhao_2022}.
\begin{align}\label{eq:z_measurement}
    z_{p_z,k} 
        = h_{p_z}(\mathbf{x}_k)+ n_P
        = \mathbf{s} {^{I_k}_G}\mathbf{R}^\top {^I_D}\mathbf{R} {^D_P}\mathbf{R} 
           ({^P}\mathbf{P}_{in} - {^P}\mathbf{P}_{k}) + n_{p_z}
\end{align}

\subsection{Visual Measurement Update}

We perform point feature tracking on a selected image and use the feature tracking results in multiple sequential images (or keyframe images) to update the system state and covariance. 
For a well-calibrated camera, the measurement of a feature in the camera frame $\{C_i\}$ is the perspective projection of its 3D position 
$\prescript{C_i}{}{\mathbf{p}}_{f} =[\prescript{C_i}{}{x},\prescript{C_i}{}{y},
\prescript{C_i}{}{z}]^\top$ onto the normalized plane, which is given by Eq.~\ref{eq:cam_measurement} where $\prescript{C_i}{}{\mathbf{p}}_{f}$ is given by Eq.~\ref{eq:pf}.
In Eq.~\ref{eq:cam_measurement} and \ref{eq:pf}, $\pi(\cdot)$ is the perspective projection function, function $\tau(\cdot)$ transforms a point based on the given transformation matrix,
$\mathbf{n}_C$ is the zero-mean white Gaussian noise for camera measurement,
$\{ \prescript{C}{I}{\mathbf{R}}, \prescript{C}{}{\mathbf{p}}_{I} \}$ is the extrinsic calibration between IMU and camera,
$\{ \prescript{I_k}{G}{\mathbf{R}}, \prescript{G}{}{\mathbf{p}}_{I_k} \}$ 
is the cloned IMU pose at image time $k$,
$\prescript{G}{}{\mathbf{p}}_{f}$ is the feature's 3D position in the global frame. 
\begin{align}\label{eq:cam_measurement}
\mathbf{z}_{C,i} 
  &= \pi(\prescript{C_i}{}{\mathbf{p}}_{f}) + \mathbf{n}_C 
   = \left[ 
        \begin{array}{c} 
           \prescript{C_i}{}{x}/\prescript{C_i}{}{z} \\
           \prescript{C_i}{}{y}/\prescript{C_i}{}{z}
        \end{array} 
     \right] + \mathbf{n}_C   
\end{align}
\vspace{-3ex}
\begin{align}\label{eq:pf}
\prescript{C_i}{}{\mathbf{p}}_{f}
  = \tau(\prescript{G}{}{\mathbf{p}}_{f},  
     \prescript{C}{I}{\mathbf{T}} \prescript{I_k}{G}{\mathbf{T}} ) 
  = \prescript{C}{I}{\mathbf{R}} \prescript{I_k}{G}{\mathbf{R}} 
     (\prescript{G}{}{\mathbf{p}}_{f} - \prescript{G}{}{\mathbf{p}}_{I_k}) 
     + \prescript{C}{}{\mathbf{p}}_{I}
\end{align}

For each feature, we compute the residual using the Jacobian matrices of the measurement function (Eq.~\ref{eq:cam_measurement}) with respect to the state and feature ($\mathbf{H}_{x}$ and $\mathbf{H}_{f}$)
\begin{align}\label{eq:res_cam_linearized}
\mathbf{\tilde{z}}_{C_k} 
  = \mathbf{H}_{x} \tilde{\mathbf{x}}_k +
    \mathbf{H}_{f} \prescript{G}{}{\tilde{\mathbf{p}}}_{f} +
    \mathbf{n}_C              
\end{align}
Next, we apply the left nullspace~\cite{MSCKF_Mourikis_2007} of the feature Jacobian to convert Eq.~\ref{eq:res_cam_linearized} to \ref{eq:res_cam_nullspace}, and used it for EKF update.
\begin{align}\label{eq:res_cam_nullspace}
\mathbf{N}^\top\mathbf{\tilde{z}}_{C_k} 
  = \mathbf{N}^\top\mathbf{H}_{x} \tilde{\mathbf{x}}_k +
    \mathbf{N}^\top \mathbf{n}_C              
\end{align}

\subsection{Keyframe Marginalization}

During underwater exploration, the vehicle may hover or move closer for detailed visual investigation.
While hovering will cause a small translation between two consecutive image frames, the vertical movements will cause a small feature disparity within two successful tracking steps.
Both cases will lead to degraded triangulation results, which brings in bad visual updates for state estimation.
In~\cite{MSCKF-Hovering_Kottas_2013}, the authors propose a marginalization strategy that updates features in a consistent way for the hovering case.
However, in a more general way, we intend to use keyframe marginalization to handle more degenerate motions.
Inspired by the keyframe-based visual SLAM~\cite{ORB-SLAM3_Campos_2021}, we have implemented a strategy to insert the IMU clones based on three criteria, feature numbers, motion constraint, and scene constraint.
If a new image has more than 50 features detected, its translation (estimated from DVL-IMU fused odometry) has exceeded 0.1m, and more than 10\% of the features from the previous frame have disappeared, this new image will be treated as a keyframe.

When we reach the maximum number of keyframes, the algorithm will perform a marginalization similar to the one in S-MSCKF~\cite{S-MSCKF_Sun_2018}. 
While the standard MSCKF \cite{MSCKF_Mourikis_2007} uses the feature measurements in several poses (one-third of slide window evenly spaced), we will only use the feature measurements in two keyframes (the oldest one and the second-latest shown in Fig.~\ref{fig:feat_marg}) for standard MSCKF-feature update to reduce the computational cost.
We also implemented another change during marginalization.
Instead of removing two poses that were used in marginalization from the state matrix, our method will keep the second-latest pose even though it was used for marginalization.
The second-latest pose is saved because of the following reasons.
Any non-max-tracked features (i.e., $f_{j+1}$ in Fig.~\ref{fig:feat_marg}) detected in second-latest pose will have minimum impact on the update because these features only have one measurement when doing the marginalization.
However, keeping these features and the pose will allow us to use them in the future for sufficient MSCKF-feature updates.

\vspace{-3ex}
\textbf{\begin{figure}[h] \centering
\includegraphics[width=0.32\textwidth,height=0.16\textwidth]{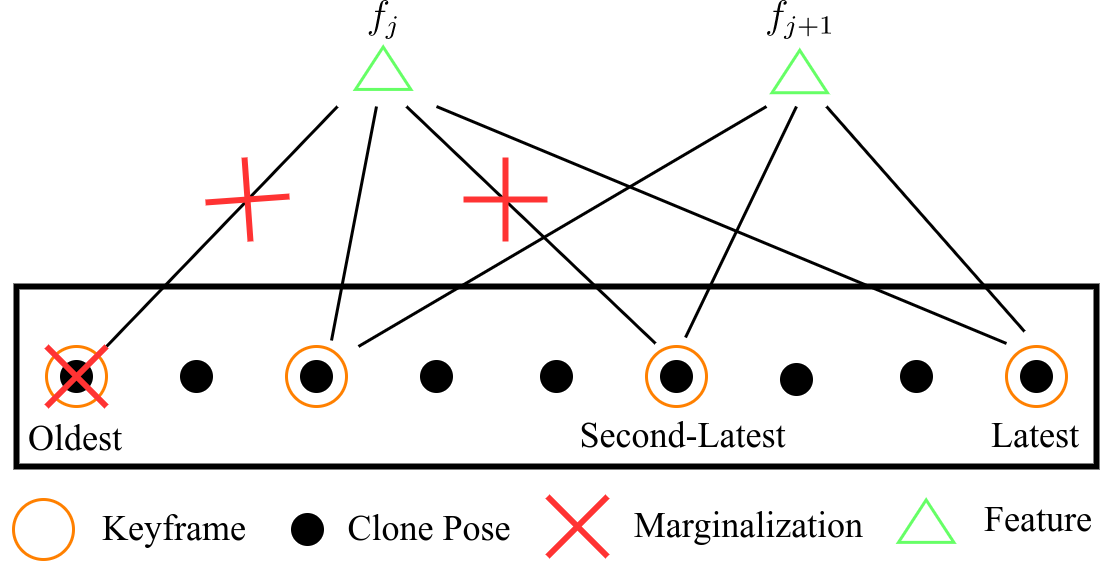}
    \caption{The feature marginalization when clones at maximum.} \label{fig:feat_marg}
\end{figure}}

%%%%%%%%%%%%%%%%%%%%%%%%%%%%%%%%%%%%%%%%%%%%%%%%%%%%%%%%%%%%%%%%%%%%%%%%%%%%%%%%
%%%%%                          Enhancement                                 %%%%%
%%%%%%%%%%%%%%%%%%%%%%%%%%%%%%%%%%%%%%%%%%%%%%%%%%%%%%%%%%%%%%%%%%%%%%%%%%%%%%%%

% !TEX root = main.tex

%%%%%%%%%%%%%%%%%%%%%%%%%%%%%%%%%%%%%%%%%%%%%%%%%%%%%%%%%%%%%%%%%%%%%%%%%%%%%%%%
%%%%%                          Enhancement                                 %%%%%
%%%%%%%%%%%%%%%%%%%%%%%%%%%%%%%%%%%%%%%%%%%%%%%%%%%%%%%%%%%%%%%%%%%%%%%%%%%%%%%%

\section{DVL-aided Feature Enhancement}
\label{sec:enhancement}
During operation, the vehicle will keep a close distance (1-2 meters) to the ice such that the up-looking camera only has a small coverage, which causes a relatively short feature tracking distance.
In other words, the relative pose changes between keyframes will be small which may lead to bad feature triangulation.
To deal with this problem, we designed a DVL-aided feature enhancement strategy that uses the DVL point clouds obtained from 4 beams to constrain the feature depth.
The detailed approach is discussed in the remaining content in this section. 

\subsection{Data Association}
Typically, 3D-2D data association between point cloud and image feature require one-by-one matching~\cite{LiDAR-SfM_Zhen_2020}. 
However, DVL only provides sparse point clouds, making exact point matching challenging.
Therefore, we attempt to find the nearest DVL point cloud around each feature, then apply bilinear interpolation to obtain the possible depth at the feature's location.
Our method has three steps, identifying the anchor frame of the feature, identifying relevant DVL point clouds, and finding the DVL point cloud associated with the camera features.
The anchor frame is selected based on the feature's location in the normalized plane. 
Among all keyframes, we select the one to be the anchor frame when the feature has a minimum offset from the camera origin because this feature will be highly likely to be bounded by the DVL points nearby in the anchor frame.
The anchor frame will be treated as the reference frame for feature triangulation. 
To identify relevant DVL point clouds, we sort the buffered DVL point cloud based on
the timestamp difference between the DVL point clouds and the anchor frame.
The $m$ number of DVL point cloud with the smallest time differences will be selected for feature enhancement.
To match the selected DVL point cloud with the feature, we design the following three-step procedure with pseudo-code displayed in Algorithm ~\ref{alg:cloud_match}.

\begin{algorithm}
  \caption{Feature and Cloud Match}\label{alg:cloud_match}
  % \small
  % \footnotesize
  \scriptsize
  % \tiny
  \Input{
        {\begin{minipage}[t]{6cm}%
             \strut   
             A list of DVL timestamps with clouds \\ 
             $ \{ \prescript{D_i}{}{t}, \prescript{D_i}{}{\mathcal{P}} \}, i \in (1,m) $ 

             A list of IMU timestamps with clones \\ 
             $ \{ \prescript{I_j}{}{t}, \prescript{G}{I_j}{T} \in \textit{SE}(3) \}, j \in (1,n) $

             The threshold of standard deviation $\sigma_z$ to filter outlier

             The normalized feature measurement $(u,v)$ and the pose of the anchor frame
             $\prescript{G}{C}{T}$
             \strut
        \end{minipage}}
        }
  \Output{cloud at camera frame $\prescript{C_i}{}{\mathcal{P}}$}

  \Let{$\prescript{C_i}{}{\mathcal{P}}$}{\textit{InitializeToEmpty}()}\;
  \ForEach{$ i \in (1,m), j \in (1,n) $}{
      \tcc{Step 1: Interpolate DVL Pose}
      \eIf{$\prescript{I_j}{}{t} \leq 
            \prescript{D_i}{}{t} \leq 
            \prescript{I_{j+1}}{}{t}$}{
          \Let{$\prescript{G}{I}{T}_{D_i}$} 
              {\textit{Interpolation}(
                  $\prescript{G}{I}{T}_{I_j}, \prescript{G}{I}{T}_{I_{j+1}},
                   \prescript{I_j}{}{t}, \prescript{I_{j+1}}{}{t}$)}
      }{
          \Continue
      }

      \tcc{Step 2: Outlier Rejection}
      \Let{$\prescript{G_i}{}{\mathcal{P}}$}
          {$\tau(\prescript{D_i}{}{\mathcal{P}}, 
                 \prescript{G}{I}{T}_{D_i} \prescript{I}{D}{T})$}\\
      \If{! \textit{Filter}($\prescript{G_i}{}{\mathcal{P}}, \sigma_z$)}{
          \Continue
      }

      \tcc{Step 3: Check Coverage}
      \Let{$\prescript{C_i}{}{\mathcal{P}}$}
          {$\mathit{\tau}(\prescript{G_i}{}{\mathcal{P}}, \prescript{G}{C}{T})$} \\
      \If{! \textit{PointInPolygon}($\mathit{\pi}(\prescript{C_i}{}{\mathcal{P}}), (u,v)$)
           }
         {\Continue}

  } 

  \Return{$\prescript{C_i}{}{\mathcal{P}}$}
\end{algorithm}
% \vspace{-2ex}
%
First, we will interpolate the IMU pose at DVL timestamp based on IMU clone poses.
For this step, We adopt the linear interpolation~\cite{Resource-Constrained-VIO_Li_2014} as described in Eq.~\ref{eq:imu_pose_interpolation},
where $\exp(\cdot)$ and $\log(\cdot)$ are the \textit{SO}(3) matrix exponential and logarithmic functions~\cite{Lie-Groups_Chirikjian_2011},  $\prescript{D}{}{t}$ is the DVL timestamp, $\prescript{I_a}{}{t}$ and 
$\prescript{I_b}{}{t}$ are the beginning and end of the IMU clone interval timestamps.
\begin{subequations}
\label{eq:imu_pose_interpolation}
\begin{align}
	\lambda 
	  &= (\prescript{D}{}{t} - \prescript{I_a}{}{t}) / (\prescript{I_b}{}{t} - \prescript{I_a}{}{t}) 
	     \label{eq:interp_lambda} \\
	\prescript{I}{G}{\mathbf{R}}
		&= \exp(\lambda \log(\prescript{I_b}{G}{\mathbf{R}} 
												 \prescript{I_a}{G}{\mathbf{R}}^\top)) 
			 \prescript{I_a}{G}{\mathbf{R}}
			 \label{eq:interp_R} \\
	\prescript{I}{G}{\mathbf{p}}
	  &=  (1-\lambda) \prescript{G}{}{\mathbf{p}_{I_a}} +
	  		\lambda \prescript{G}{}{\mathbf{p}_{I_b}}
	  		\label{eq:interp_p} 
\end{align}
\end{subequations}

Second, we remove the outliers in the selected DVL point clouds.
The DVL point cloud $\prescript{D_i}{}{\mathcal{P}}$ will be transformed to the global frame based on the known transform between IMU and DVL. 
Since the vehicle will be operated close to the ice (1-2 meters), the 4 points obtained in a single DVL measurement will be close to each other, therefore, the terrain will not change significantly.
However, during field trials, we may encounter ice-openings (either manually drilled or naturally formed), which causes outliers in the DVL point cloud.
Therefore, we remove the selected DVL point cloud with depth outside the standard deviation, $\sigma_z$.

Third, we will check each selected DVL point cloud (4 points from each beam) if it bounds the feature. 
The DVL point cloud will be projected to the normalized plane of the anchor frame. 
If the feature is located inside the area bounded by 4 DVL points in the normalized plane, we will then use the DVL point cloud for feature enhancement.

\subsection{Feature Enhancement}

%%% depth interpolation
In~\cite{TerrainAidedNav-Glider_Claus-2015}, bilinear interpolation is applied to obtain the specific terrain depth from a digital elevation model (DEM).
However, bilinear interpolation requires data located inside a rectilinear grid, which is not our case since the projected DVL point cloud could be an arbitrary quadrilateral. 
To transform an arbitrary quadrilateral into a unit square, we adopt a bilinear mapping function~\cite{FEM_Hughes_2000}:
\begin{subequations}
\label{eq:bilinear_mapping}
\begin{align}
  u(\xi,\eta) 
    &= \alpha_0 + \alpha_1\xi + \alpha_2\eta + \alpha_3\xi\eta
       \label{eq:mapping_x} \\ 
  v(\xi,\eta) 
    &= \beta_0 + \beta_1\xi + \beta_2\eta + \beta_3\xi\eta
       \label{eq:mapping_y}        
\end{align}
\end{subequations} 
\vspace{-5ex}
\begin{figure}[ht!]
    \centering
    \includegraphics[width=0.9\linewidth]{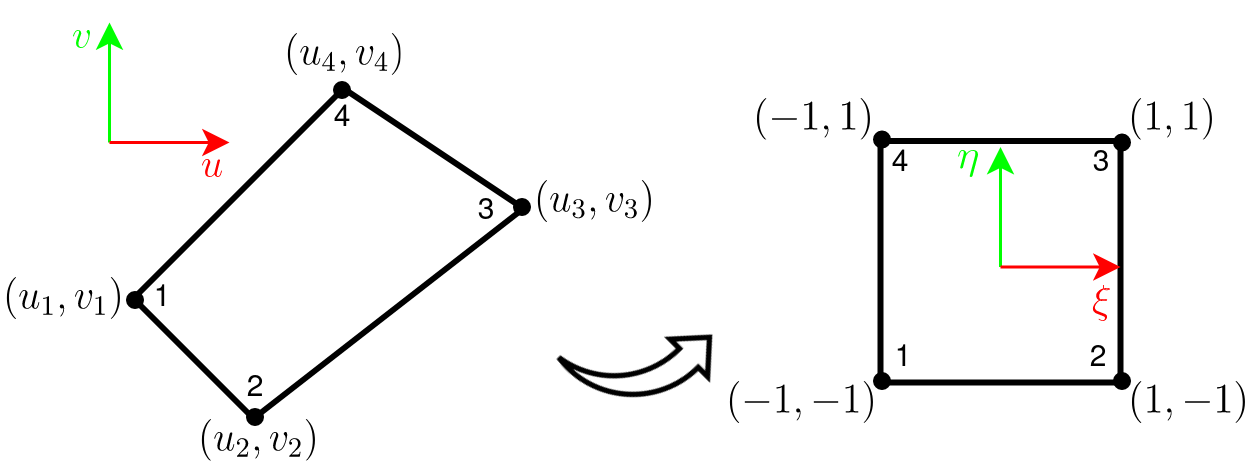}
    \caption{The mapping process from an arbitrary quadrilateral to unit square}
    \label{fig:bilinear_interpolation}
\end{figure}

%\vspace{-1ex}

\begin{figure}[h] \centering
    \includegraphics[width=0.45\textwidth]{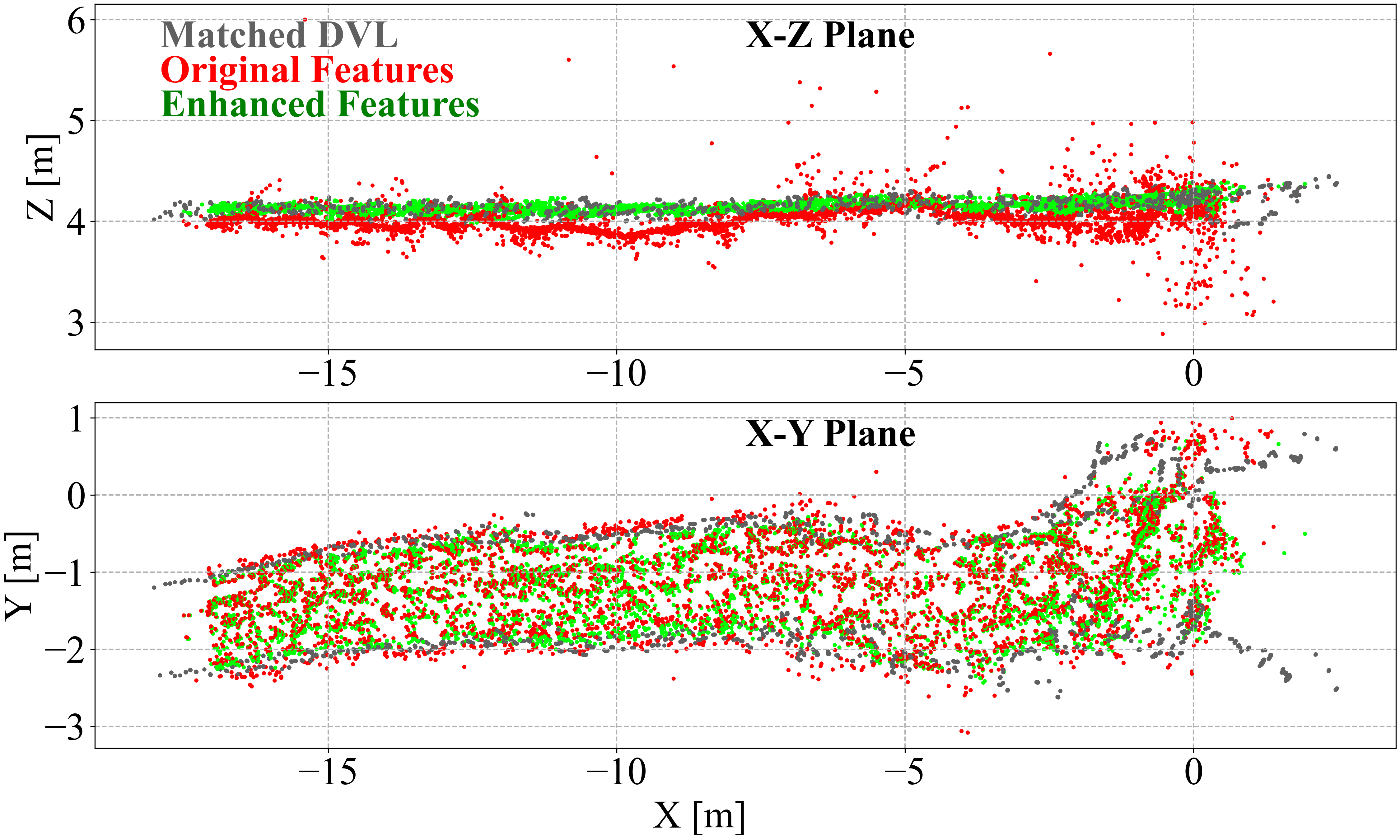}
    \\
    \includegraphics[width=0.45\textwidth]{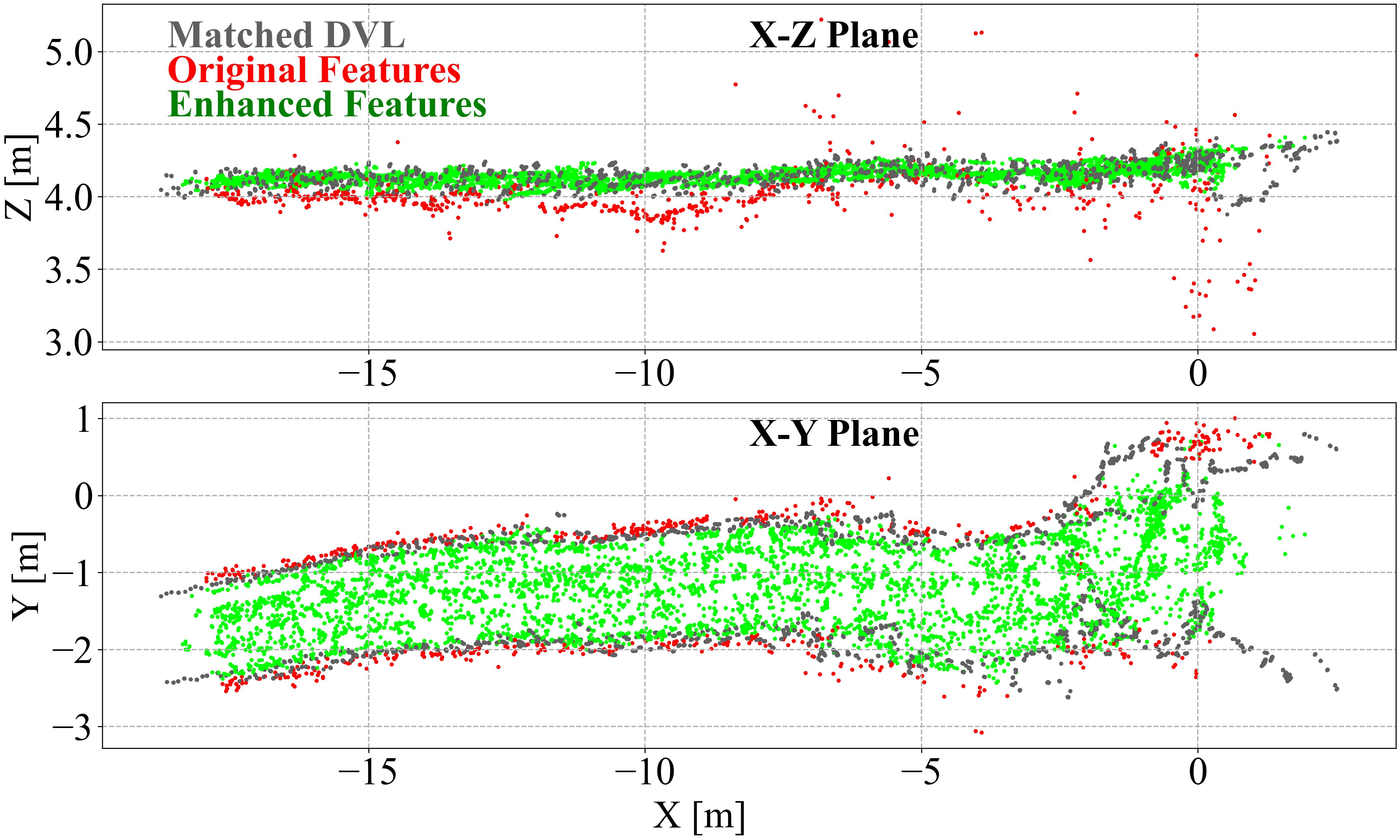}
    \\
    \caption{Top: Comparison between triangulation and enhanced feature position,
         Bottom: Comparison between matched and not matched features. Matched DVL point cloud (only placed on edges) is gray, normal triangulated features are red, and enhanced 
         features are green.} 
    \label{fig:depth_enhance}
\end{figure}
\vspace{-3ex}

The coefficients $\boldsymbol{\alpha}, \boldsymbol{\beta}$ can be solved after substituting the normalized cloud point coordinates $\{\mathbf{u}_D,\mathbf{v}_D\}$ and square vertex coordinates $\{\boldsymbol{\xi}, \boldsymbol{\eta}\}$ shown in Figure~\ref{fig:bilinear_interpolation}. 
After that, the normalized feature coordinate can be mapped into the unit square and the bilinear interpolation can be applied to estimate the feature depth. 

%%% position recovery
%
Feature position recovery has two steps, normal triangulation and position enhancement.
First, for each feature, Direct linear transformation (DLT)~\cite{MVG-CV_Hartley_2004} triangulation is applied if more than two observations in the keyframes. 
After that, inverse-depth parameterization~\cite{MSCKF_Mourikis_2007} based nonlinear optimization is used for refining the feature position. 
Next, the feature position obtained from the two previous steps is corrected by multiplying the scaling ratio ($z_a/z_b$) where $z_b$ is the feature depth computed from the two previous steps and $z_a$ is the feature depth interpolated from the DVL point cloud.
This change will then be incorporated into the visual measurement update discussed in Section III.E.

As shown in Figure~\ref{fig:depth_enhance},  the first and second figures visualize the normal triangulation result (red) and enhanced feature position (green). 
The feature depth is well recovered since they almost align with the DVL point clouds (gray). The third and fourth figures visualize the enhanced (green) and un-enhanced (red) features. 
Most of the enhanced features are located inside DVL point clouds.
Because of the keyframe strategy, there is no sufficient measurement update for each image frame. We intend to keep both enhanced and not enhanced features for measurement update.

%%%%%%%%%%%%%%%%%%%%%%%%%%%%%%%%%%%%%%%%%%%%%%%%%%%%%%%%%%%%%%%%%%%%%%%%%%%%%%%%
%%%%%                           Experiment Results                         %%%%%
%%%%%%%%%%%%%%%%%%%%%%%%%%%%%%%%%%%%%%%%%%%%%%%%%%%%%%%%%%%%%%%%%%%%%%%%%%%%%%%%

% !TEX root = main.tex

%%%%%%%%%%%%%%%%%%%%%%%%%%%%%%%%%%%%%%%%%%%%%%%%%%%%%%%%%%%%%%%%%%%%%%%%%%%%%%%%
%%%%%                           		Experiment                         		 %%%%%
%%%%%%%%%%%%%%%%%%%%%%%%%%%%%%%%%%%%%%%%%%%%%%%%%%%%%%%%%%%%%%%%%%%%%%%%%%%%%%%%

\begin{figure}[h]
    \centering
    \includegraphics[scale=0.35]{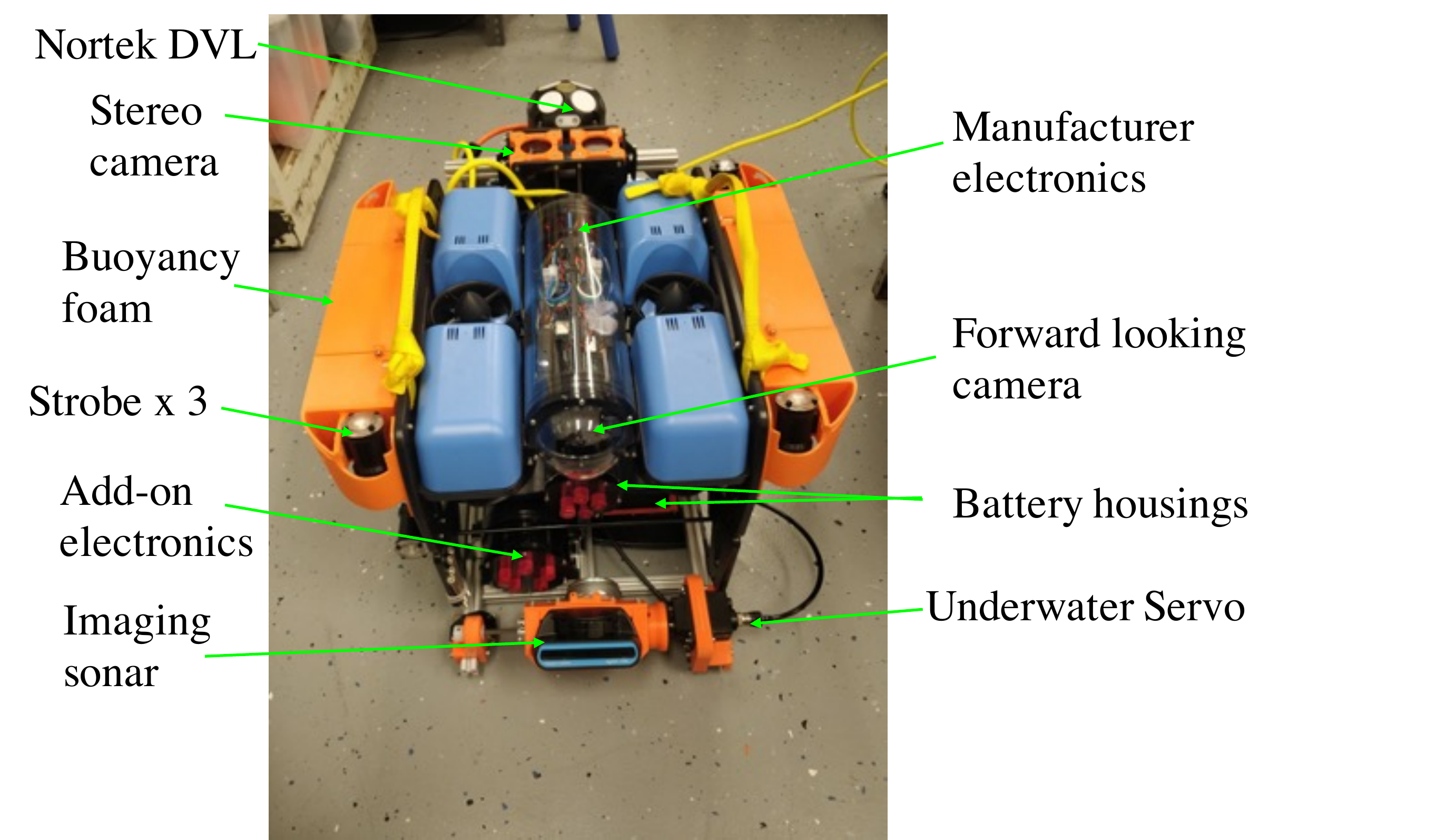}
    \caption{The modified BlueROV-2 used in the  experiment.}
    \label{fig:rov}
\end{figure}
% \vspace{-2ex}

\section{Experiment Results}
\label{sec:experiment}

\begin{figure*}[t] \centering
 \vspace{-5ex}
 \includegraphics[width=0.99\textwidth,height=0.2\textwidth]{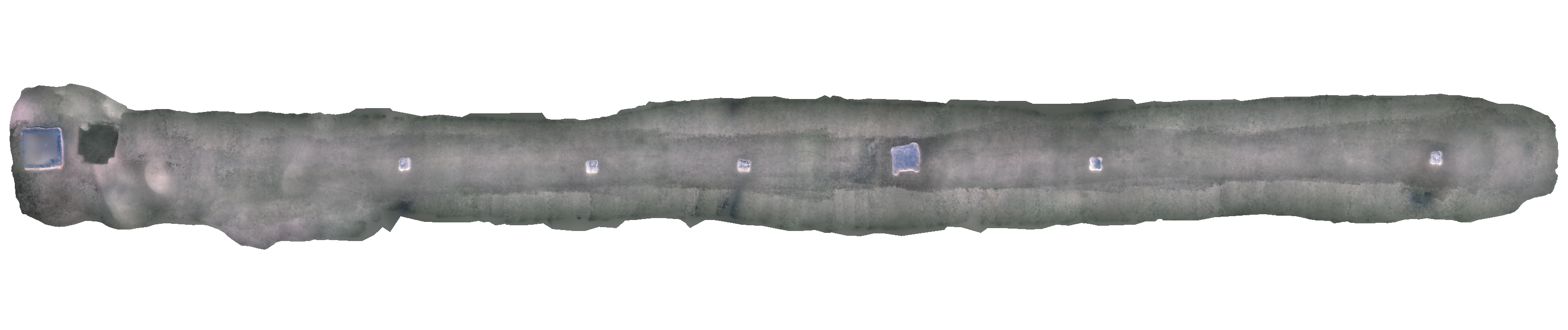}
    \vspace{-6ex}
    \caption{The Metashape reconstructed result. The largest ice-hole on the left side is the starting point for the vehicle, the total length of this reconstructed result is roughly 40 meters} \label{fig:metashape}
\end{figure*}

\subsection{Experiment data set}
% hardware 
In March 2021, we have conducted an under-ice experiment under the frozen Keweenaw Waterway in Michigan using a modified BlueROV2~\cite{UnderIce-ROV_Zhao_2021} with a suite of sensors shown in Fig.~\ref{fig:rov}. 
The ice thickness is about 30 cm.
An ice hole (about 1m by 1m) was cut for deploying the ROV while several small ice holes (shown in Fig. \ref{fig:metashape}) were also drilled along the transect. They are spaced at 10 meters except the second one.
During the experiment, the ROV is remotely controlled by the pilot to drive along a straight line multiple times (roughly 40 meters each way) from a position at 4 meters deep, resulting in a total traveling distance of about 200 m and a total duration of about 20 minutes.

We used the experimental data set to validate our proposed sensor fusion framework.
In the data set, the up-looking stereo-camera is running at 15 Hz with a raw image size of 1616 by 1240 pixels. 
Only one camera was used for this experiment. The upward-looking DVL is pinging at 4 Hz, the IMU is running at 100 Hz, and the pressure sensor on the DVL is sampling at 2 Hz.
The standard deviation (SD) of the DVL single ping at 3 m/s is about 0.005 m/s from Norteck technical specification. During the data collection, the vehicle moves at about 0.4 m/s and the transformation between DVL and IMU is roughly measured. Therefore, we set the velocity SD to between 0.0375 - 0.1 m/s for the experiment.
The original result shown in \cite{UnderIce-ROV_Zhao_2021} used the robot localization without correcting the time delays (about 10 seconds) between the IMU and DVL due to the DVL driver issue.
Even though the localization in \cite{UnderIce-ROV_Zhao_2021} shows a low drift, it may be a coincidence.
In this data set, we have corrected the delays during the validation process.
One unique feature in this data set is that, occasionally, the ROV is controlled to hover in place.
Such maneuvers will challenge the visual SLAM performance since during the hovering no significant translation is available for feature triangulation. 
In application, hovering may be needed in several key locations during an under-ice exploration to collect more measurements on abnormal biogeochemical processes, e.g., a salt brine injection and algae bloom.

\subsection{Results}
The ground truth vehicle path is generated using Agisoft Metashape based on SfM technique, a rendering of the ice surface is shown in Fig.~\ref{fig:metashape}.
For comparison, we use the evo~\cite{evo_Grupp_2017} toolbox to align (recovery orientation and scale) Metashape ground truth and estimated odometry created from different sensor fusion methods.
We only selected a short amount of time (90 seconds about 10 meters) at the beginning for alignment.
Herein, we compare the localization results from 10 settings, as shown in Table~\ref{tab:odom_setting} against the ground truth path.
\begin{table}[h]
    \caption{Setup with different sensor suites and features. "Y" means used and "N" means not used.}
    \label{tab:odom_setting}
    \centering
    \begin{tabular}{c|c|c|c|c|c|c|c|c|c|c}
    \hline
    \textbf{Case \#} &\textbf{1} &\textbf{2} &\textbf{3} &\textbf{4} &\textbf{5} 
                     &\textbf{6} &\textbf{7} &\textbf{8} &\textbf{9} &\textbf{10} \\
    \hline
    Visual      &Y  &Y  &Y  &Y  &Y  &N  &\textit{Y}  &\textit{Y}  &\textit{Y}  &\textit{Y}  \\    
    \hline
    DVL         &Y  &Y  &Y  &Y  &Y  &Y  &\textit{N}  &\textit{N}  &\textit{N}  &\textit{N}  \\
    \hline
    IMU         &Y  &Y  &Y  &Y  &Y  &Y  &\textit{Y}  &\textit{Y}  &\textit{Y}  &\textit{Y}  \\
    \hline
    Pressure    &Y  &Y  &Y  &Y  &N  &Y  &\textit{N}  &\textit{N}  &\textit{Y}  &\textit{Y}  \\
    \hline
    Enhancement &Y  &N  &Y  &N  &Y  &N  &\textit{N}  &\textit{N}  &\textit{N}  &\textit{N}  \\
    \hline
    Keyframe    &Y  &Y  &N  &N  &Y  &N  &\textit{Y}  &\textit{N}  &\textit{Y}  &\textit{N}  \\
    \hline
    \end{tabular}
\end{table} 
% \vspace{-2ex}

\begin{figure*}[t] \centering
 % \vspace{-1ex}
    \makebox[0.49\textwidth]{\small (a) }
    \makebox[0.48\textwidth]{\small (b) }
    % \vspace{+2mm}
    % \\
    \includegraphics[width=0.48\textwidth]{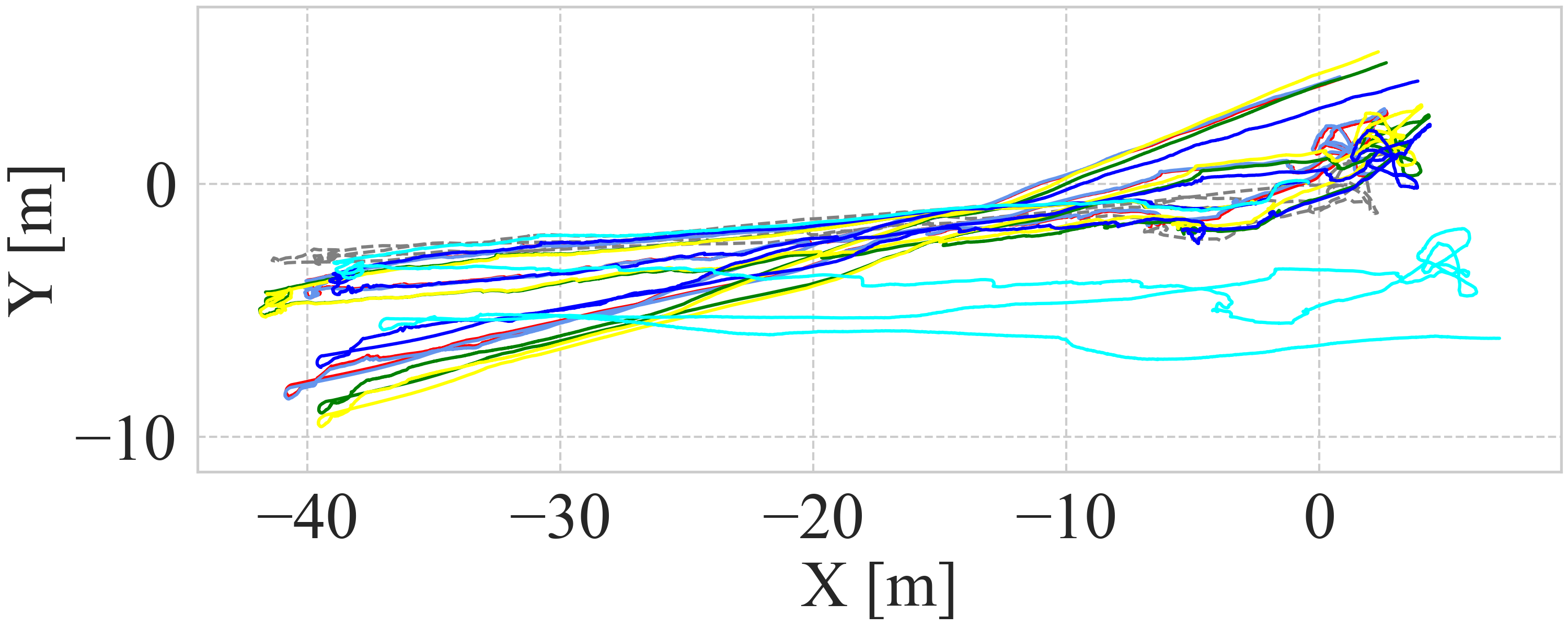} 
    \hspace{+2mm}
    \includegraphics[width=0.48\textwidth]{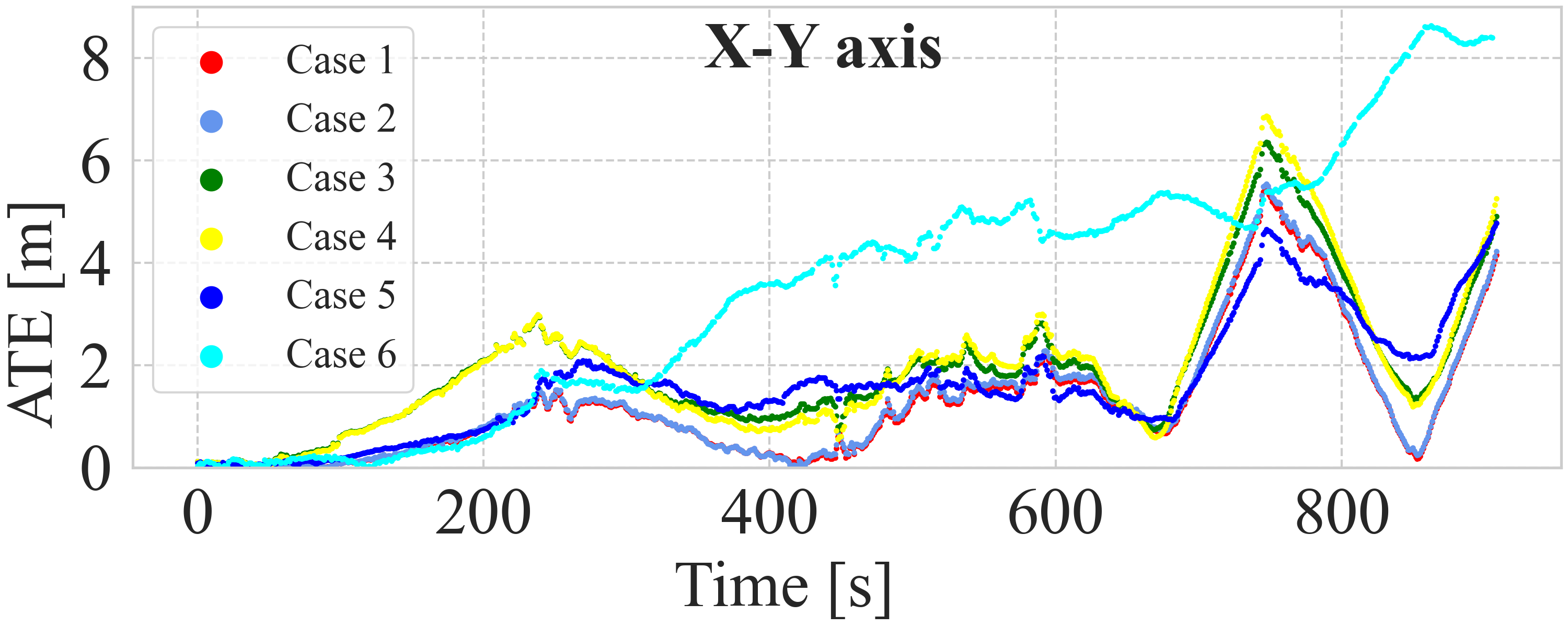}
    \caption{The evaluation result. (a) The aligned trajectories for case 1 to case 6. (b) The ATE translation RMSE for X-Y axis. } 
    \label{fig:plot_result}
\end{figure*}

For all tracks, we used identical parameters in the MSCKF and system initialization is conducted using the method from~\cite{INS-DVL-Pressure_Zhao_2022}. 
We used CLAHE~\cite{CLAHE_Pizer_1987} with KLT~\cite{KLT_Lucas_1981} method for the front-end feature tracking because the descriptor based methods, such as the ORB and KAZE, didn't provide us with a consistent tracking result. 
We found that the descriptor-based method can be confused by the air bubbles in the ice which appear in similar shapes and sizes.
We present all the resulting vehicle paths estimated from case $1\thicksim6$ in Fig.~\ref{fig:plot_result}(a) with different colors. Noted that VIO options (case $7\thicksim10$) are not visualized since those runs failed quickly at the beginning because of the hovering maneuvers.  
From Fig.~\ref{fig:plot_result}(a), we can easily observe that the odometry generated without the visual assist (case 6) is drifting away from the ground truth. 
In contrast, the paths generated with visual assistant stay closer to the ground truth path, especially, during the first and the second transects.
We believe that the angle offset between tracks and the ground truth during the third and fourth segments may due to the hovering maneuverings (2-3 minutes) near the ROV deployment hole at the end of the second segment. 

% \vspace{-2ex}
% \textbf{\begin{figure}[h] \centering
% \includegraphics[width=0.5\textwidth,height=0.3\textwidth]{images/aligned_traj.png}
%     \caption{The resulting trajectories for all the cases.} \label{fig:aligned_traj}
% \end{figure}}
% \vspace{-2ex}

To further compare the performance in different cases, we computed the Absolute Trajectory Error (ATE)~\cite{VIO-EVO_Zhang_2018} in X and Y and the X-Y plane between the ground truth path and aligned each odometry.
The statistical values are listed in Table~\ref{tab:rmse}
and the X-Y errors are presented in Fig.~\ref{fig:plot_result}(b).
Based on that, we could see that the integration of visual measurement into the MSCKF will help with reducing errors. 
Overall, the drift in the Y direction (transversal to the vehicle transects) is higher than in the x direction. 
This may be the fact that the vehicle's sway velocity is slightly small than its surge speed.
Therefore, a lower SNR may be expected in the transversal direction, causing the increased drift.

\begin{table}[h!]
  \begin{center}
    \caption{ATE with RMSE metric for different cases.}
    \label{tab:rmse}
    \begin{tabular}{c|c|c|c|c|c|c} 
      \hline
      \textbf{Case \#} &\textbf{1} &\textbf{2} &\textbf{3} &\textbf{4} &\textbf{5} &\textbf{6}\\
      \hline
      RMSE(X)   &\textbf{0.39}  &0.42  &1.34  &1.23   &1.45  &3.54  \\
      RMSE(Y)   &\textbf{1.82}  &1.86  &2.09  &2.29   &1.45  &2.87  \\
      RMSE(X-Y) &\textbf{1.11}  &1.14  &1.71  &1.76   &1.45  &3.21  \\
      \hline
    \end{tabular}
  \end{center}
\end{table}
\vspace{-1ex}

When comparing the statistical values in Table~\ref{tab:rmse}, we have several findings.
First, visual-fused solutions are better than case 6 which only used the DVL, IMU, and pressure measurements.
Second, case 1 and 2 are better than case 3 and 4. 
This comparison allows us to highlight the benefit of having keyframe selection mechanism which allows a longer translation for a better result in feature triangulation, ultimately affecting the localization.
Third, case 1 is better than case 5 means pressure update actually helped the 2D pose estimation.
Fourth, our method (case 1 with DVL-aided feature enhancement and keyframe selection enabled) produced the lowest RMSE. 
However, case 2 (with DVL-aided feature enhancement disabled but keyframe selection enabled) is only slightly worse than case 1.
This small improvement may be mainly caused by two following reasons.
First, the detected visual features are too close to the vehicle (roughly between 1-2 meters). 
Therefore, the feature's position in z corrected by the DVL measurements is relatively small (even though the improvement is visible in Fig.~\ref{fig:depth_enhance}), resulting in a small impact on the state estimation.
Second, the feature measurements noise is set to 0.09 pixel which is relatively high compared to 0.0035 we set when testing the VIO on simulated data from OpenVINS.
We also tried 0.01 for our data, the localization error was larger than the shown result.
Therefore, we think there are still room for improvement, especially, in the front-end feature tracking.

% \begin{figure}[h]
%     \centering
%     \includegraphics[scale=0.35]{images/error.png}
%     \caption{The errors in X-axis, Y-axis and X-Y plane.}
%     \label{fig:errors}
% \end{figure}
% \vspace{-2ex}

% init alignment RMSE:
%   - DIPO:  0.04372366500998966
%   - VDIPO: 0.042506709947047125
%   - VDIPO-wo-enhance: 0.041838946871800786
%   - VDIPO-wo-enhance-keyframe: 0.04293661481755104
%   - VDIPO-wo-keyframe: 0.04353596573555296 

%%%%%%%%%%%%%%%%%%%%%%%%%%%%%%%%%%%%%%%%%%%%%%%%%%%%%%%%%%%%%%%%%%%%%%%%%%%%%%%%
%%%%%                     Conclusions and Future Work                      %%%%%
%%%%%%%%%%%%%%%%%%%%%%%%%%%%%%%%%%%%%%%%%%%%%%%%%%%%%%%%%%%%%%%%%%%%%%%%%%%%%%%%

% !TEX root = main.tex

%%%%%%%%%%%%%%%%%%%%%%%%%%%%%%%%%%%%%%%%%%%%%%%%%%%%%%%%%%%%%%%%%%%%%%%%%%%%%%%%
%%%%%                     				 Conclusions                      			 %%%%%
%%%%%%%%%%%%%%%%%%%%%%%%%%%%%%%%%%%%%%%%%%%%%%%%%%%%%%%%%%%%%%%%%%%%%%%%%%%%%%%%
% \vspace{-2mm}
\section{Conclusions and Future Plan}
\label{sec:conclusions}

In this paper, we presented a tightly-couple Visual-DVL-Inertial odometry for underwater robots.
A modified keyframe selection and marginalization method was introduced, and a DVL-aided feature enhancement approach is realized to further improve the localization performance.
With those key contributions, we have validated the complete framework with a challenging under-ice data set.
Based on the statistical values, we found our method has the lowest error with an RMSE of 1.11 meters for X-Y plane translation. 

We are planning our future research in two directions, upgrading the existing frame and integrating more perception sensors.
Currently, our visual measurement noise is set relatively high.
But, we expect the improved front-end feature tracking could allow us to lower the measurement noise for better localization results.
Previous research~\cite{DVL-IMU-Calib_Troni_2015} has shown that a well-calibrated transform between DVL and IMU will improve navigation accuracy.  
We are also interested in integrating extra perception sensors such as a forward-looking sonar that could provide feature measurements at a further distance with scales.
However, imaging sonar normally has a wide elevation angle that could not be directly resolved from the image, which will pose challenges in feature detection and tracking.
In the future, we are also interested in evaluating our algorithm on other underwater datasets and comparing it with the state-of-art SLAM such as SVIn2~\cite{SVIn2_Rahman_2019} and OpenVINS~\cite{OpenVINS_Geneva_2020}.

%%%%%%%%%%%%%%%%%%%%%%%%%%%%%%%%%%%%%%%%%%%%%%%%%%%%%%%%%%%%%%%%%%%%%%%%%%%%%%%%
% This command serves to balance the column lengths
% on the last page of the document manually. It shortens
% the textheight of the last page by a suitable amount.
% This command does not take effect until the next page
% so it should come on the page before the last. Make
% sure that you do not shorten the textheight too much.
% \addtolength{\textheight}{-12cm}

%%%%%%%%%%%%%%%%%%%%%%%%%%%%%%%%%%%%%%%%%%%%%%%%%%%%%%%%%%%%%%%%%%%%%%%%%%%%%%%%
%%%%%                              Appendix                                %%%%%
%%%%%%%%%%%%%%%%%%%%%%%%%%%%%%%%%%%%%%%%%%%%%%%%%%%%%%%%%%%%%%%%%%%%%%%%%%%%%%%%

% \input{8_appendix}

%%%%%%%%%%%%%%%%%%%%%%%%%%%%%%%%%%%%%%%%%%%%%%%%%%%%%%%%%%%%%%%%%%%%%%%%%%%%%%%%
%%%%%                         Acknowlegment                                %%%%%
%%%%%%%%%%%%%%%%%%%%%%%%%%%%%%%%%%%%%%%%%%%%%%%%%%%%%%%%%%%%%%%%%%%%%%%%%%%%%%%%

% \input{9_acknowlegment}

%%%%%%%%%%%%%%%%%%%%%%%%%%%%%%%%%%%%%%%%%%%%%%%%%%%%%%%%%%%%%%%%%%%%%%%%%%%%%%%%
%%%%%                             Reference                                %%%%%
%%%%%%%%%%%%%%%%%%%%%%%%%%%%%%%%%%%%%%%%%%%%%%%%%%%%%%%%%%%%%%%%%%%%%%%%%%%%%%%%

%% This put reference into a new package
% \clearpage

\bibliographystyle{IEEEtran}
\bibliography{IEEEabrv,reference}

\end{document}